\documentclass[10pt, conference, compsocconf]{IEEEtran}

\ifCLASSINFOpdf
\else
\fi
\usepackage{amsmath}
\usepackage{amssymb}
\usepackage{times}
\usepackage{multicol}
\usepackage{multirow}
\usepackage{bbding}
\usepackage{booktabs}
\usepackage{algorithm}
\usepackage{algorithmic}
\usepackage{graphicx}
\usepackage{epstopdf}
\usepackage{subfigure}
\usepackage{color}
\usepackage{xcolor}
\usepackage{hyperref}
\usepackage{url}
\usepackage{array}
\usepackage{bm}
\usepackage{enumitem}
\begin{document}
	\title{Stratified Transfer Learning for\\
		Cross-domain Activity Recognition}

\author{\IEEEauthorblockN{
		Jindong Wang\IEEEauthorrefmark{1}\IEEEauthorrefmark{2},
		Yiqiang Chen\IEEEauthorrefmark{1}\IEEEauthorrefmark{2},
		Lisha Hu\IEEEauthorrefmark{3},
		Xiaohui Peng\IEEEauthorrefmark{1}\IEEEauthorrefmark{2},
		Philip S. Yu\IEEEauthorrefmark{4}
	}
	\IEEEauthorblockA{\IEEEauthorrefmark{1}Beijing Key Laboratory of Mobile Computing and Pervasive Device\\Institute of Computing Technology, Chinese Academy of Sciences, Beijing, China}
	\IEEEauthorblockA{\IEEEauthorrefmark{2}University of Chinese Academy of Sciences, Beijing, China}
	\IEEEauthorblockA{\IEEEauthorrefmark{3}Institute of Information Technology, Hebei University of Economics and Business, Shijiazhuang, China}
	\IEEEauthorblockA{\IEEEauthorrefmark{4}Department of Computer Science, University of Illinois at Chicago, IL, USA}
	Email:\{wangjindong,yqchen,pengxiaohui\}@ict.ac.cn, hulisha@heuet.edu.cn, psyu@uic.edu}
	
\maketitle

\begin{abstract}
\label{abstract}
In activity recognition, it is often expensive and time-consuming to acquire sufficient activity labels. To solve this problem, transfer learning leverages the labeled samples from the source domain to annotate the target domain which has few or none labels. Existing approaches typically consider learning a global domain shift while ignoring the intra-affinity between classes, which will hinder the performance of the algorithms. In this paper, we propose a novel and general cross-domain learning framework that can exploit the intra-affinity of classes to perform intra-class knowledge transfer. The proposed framework, referred to as \underline{S}tratified \underline{T}ransfer \underline{L}earning~(STL), can dramatically improve the classification accuracy for cross-domain activity recognition. Specifically, STL first obtains pseudo labels for the target domain via majority voting technique. Then, it performs intra-class knowledge transfer iteratively to transform both domains into the same subspaces. Finally, the labels of target domain are obtained via the second annotation. To evaluate the performance of STL, we conduct comprehensive experiments on three large public activity recognition datasets~(i.e. OPPORTUNITY, PAMAP2, and UCI DSADS), which demonstrates that STL significantly outperforms other state-of-the-art methods w.r.t. classification accuracy~(improvement of 7.68\%). Furthermore, we extensively investigate the performance of STL across different degrees of similarities and activity levels between domains. And we also discuss the potential of STL in other pervasive computing applications to provide empirical experience for future research.
\end{abstract}

\IEEEpeerreviewmaketitle

\section{Introduction}
Human activity recognition~(HAR) is a hot research topic in pervasive computing. HAR has been widely applied to many applications such as indoor localization~\cite{xu2016indoor}, sleep state detection~\cite{zhao2017sleep}, and smart home sensing~\cite{wen2016adaptive}. The key to successful HAR is to build accurate and robust models using sufficient labeled activity data. Unfortunately, it is often expensive and time-consuming to obtain enough labeled data. In this case, it is necessary to perform \textit{cross-domain activity recognition}~(CDAR) by exploiting the labeled activity data from one auxiliary domain to annotate the unlabeled activities in another domain.

\begin{figure}[t!]
	\centering
	\includegraphics[scale=0.4]{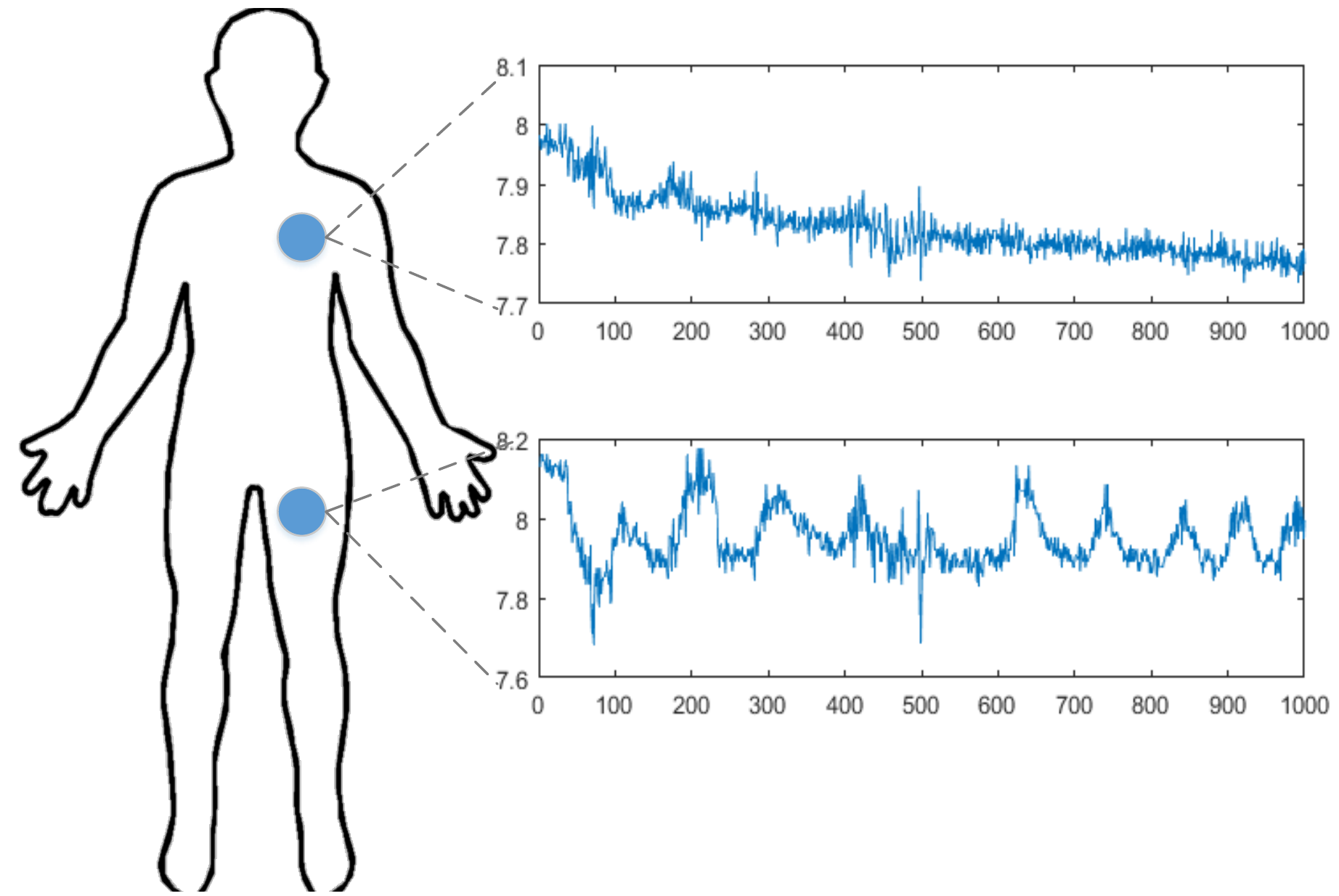}
	\vspace{-.1in}
	\caption{Two accelerometer readings of the same device when a person is walking. It is obvious that those two readings follow different distributions.}
	\label{fig-position}
	\vspace{-.2in}
\end{figure}

An intuitive example of CDAR is given in Fig.~\ref{fig-position}, which shows two accelerometer readings on the chest~(\textit{C}) and the leg~(\textit{L}) from one person. Such scenarios happen very often: If we only have the activity data and labels on \textit{C} but labels on \textit{L} are missing or wrongly labeled, we can certainly use the data from \textit{C} to recognize the activity on \textit{L}. Because the signals of \textit{C} and \textit{L} can be related according to the body structure. Then the challenge arises: It can be seen that the two sensor readings are under two totally different distributions from Fig.~\ref{fig-position}. Therefore, it is challenging to design algorithms to tackle CDAR problem. 

Several dimensionality reduction methods have been proposed to resolve CDAR, such as principal component analysis~(PCA), locally linear embedding~(LLE), and kernel principal component analysis~(KPCA)~\cite{fodor2002survey}. Dimensionality reduction does not require domain knowledge, but it ignores the divergence between domains. To this end, transfer learning~\cite{pan2010survey} has made considerable progress for cross-domain tasks by leveraging the labeled samples from other auxiliary domains. The key to transfer learning is to utilize the \textit{similarity} between those different but related domains. Existing transfer learning approaches disentangled the cross-domain learning problems either by exploiting the correlations between features~\cite{blitzer2006domain,kouw2016feature}, or transforming both the source and target domains into a new shared feature space~ \cite{pan2011domain,gong2012geodesic,long2015domain}. 

\begin{figure}[t!]
	\centering
	\subfigure[Original target]{
		\centering
		\includegraphics[scale=0.42]{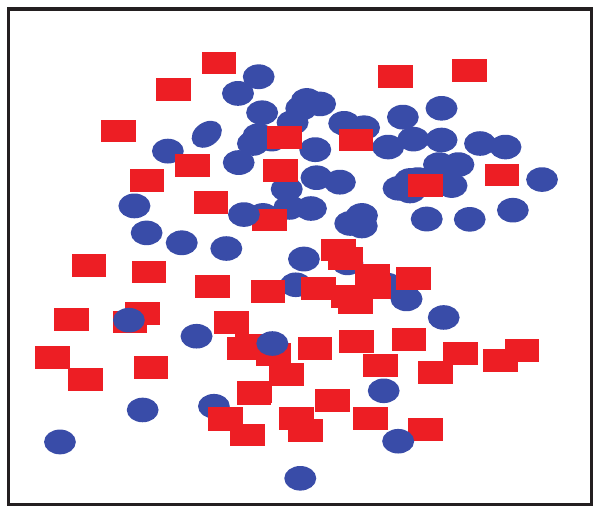}
		\label{fig-sub-target}}
	\subfigure[Learn global shift]{
		\centering
		\includegraphics[scale=0.42]{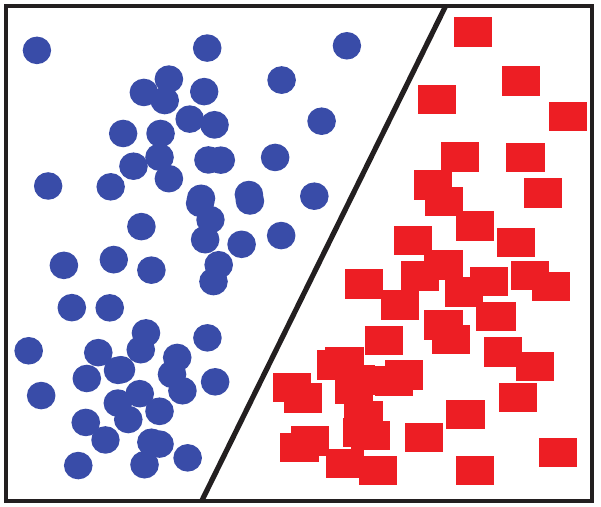}
		\label{fig-sub-global}}
	\subfigure[Exploit intra-affinity]{
		\centering
		\includegraphics[scale=0.42]{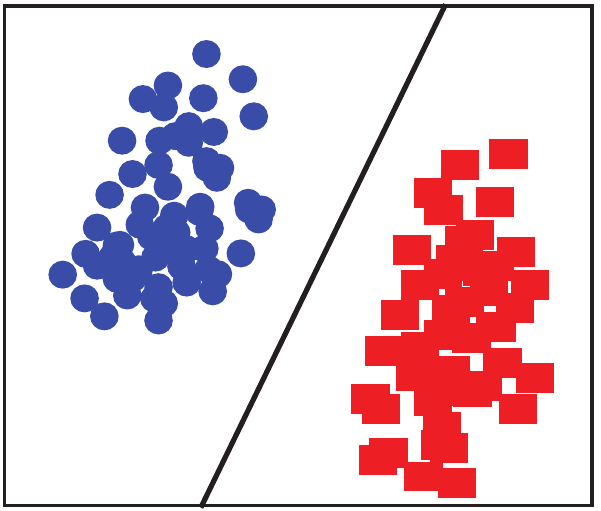}
		\label{fig-sub-stl}}
	\label{fig-intro}
	\vspace{-.1in}
	\caption{Illustration of learning global domain shift~(existing approaches) and exploiting intra-affinity~(STL) for the same original target domain.}
	\vspace{-.25in}
\end{figure}

Those approaches tend to learn a global domain shift by projecting all samples in both domains into a single subspace. However, they fail to consider the intra-affinity within classes~\cite{lin2016cross}. According to Fig.~\ref{fig-intro}, learning the global domain shift~(Fig.~\ref{fig-sub-global}) can only learn a general hyperplane between classes, and this hyperplane is loose; while by exploiting intra-class affinity~(Fig.~\ref{fig-sub-stl}), classes can further be clustered more tightly. Therefore, it is necessary to exploit the intra-affinity of classes to overcome the limitation of global domain shift.

According to~\cite{elhamifar2013sparse}, the data samples from the same class should lay on the same subspace, even if they belong to different domains. By using this assumption, in this paper, we propose a novel and general Stratified Transfer Learning~(STL) framework. STL can simultaneously transform the same classes of the source and target domains into the same subspaces. Compared to existing methods, STL exploits the intra-affinity of classes to perform intra-class knowledge transfer. Thus, the global domain shift can be alleviated. Concretely speaking, STL first obtains pseudo labels for the target domain via a majority voting technique. Then, it performs intra-class knowledge transfer to transform the source and target domain into the same subspaces. Finally, the labels of target domain are obtained through a second annotation step. Comprehensive experiments on three large public activity recognition datasets (i.e. OPPORTUNITY, PAMAP2, and UCI DSADS) demonstrate that STL outperforms other five state-of-the-art methods with a significant improvement of \textbf{7.68}\% w.r.t. classification accuracy.

\textbf{Contributions.} Our contributions are mainly three-fold:

1) We propose a novel and general cross-domain learning framework STL. Different from existing methods that only obtain a global domain shift, STL is capable of exploiting the intra-affinity of classes to perform intra-class knowledge transfer. STL is a general framework and can be tailored according to specific applications.

2) We conduct comprehensive experiments for cross-domain activity recognition on three public datasets~(i.e. OPPORTUNITY, PAMAP2, and UCI DSADS) to evaluate STL. Experiments demonstrate the superiority of STL over other state-of-the-art methods w.r.t. classification accuracy.

3) We extensively investigate the performance of STL on different degrees of task similarity and different levels of activities. And we additionally explore the potential of STL in other pervasive computing applications, providing experience for future research.

The rest of this paper is organized as follows. In Section~\ref{sec-related}, we review the related work. Section~\ref{sec-stl} introduces the proposed STL framework. In Section~\ref{sec-exp} we present experimental evaluation and analysis of STL and other comparison methods. In Section~\ref{sec-dis}, we discuss the potential of STL in other real applications. Finally, the conclusion and future work are presented in Section~\ref{sec-conclu}.

\section{Related Work}
\label{sec-related}
Our work is mainly related to activity recognition and transfer learning. In this section, we will discuss those two areas and their intersections.

\subsection{Activity Recognition}
Human Activity recognition has been a popular research topic in pervasive computing~\cite{bulling2014tutorial} for its competence in learning profound high-level knowledge about human activity from raw sensor inputs. Several survey articles have elaborated the recent advance of activity recognition using conventional machine learning~\cite{bulling2014tutorial,lara2013survey} and deep learning~\cite{wang2017deep} approaches.

Conventional machine learning approaches have made tremendous progress on HAR by adopting machine learning algorithms such as similarity-based approach~\cite{zheng2011user,chen2016ocean}, active learning~\cite{hossain2016active}, crowdsourcing~\cite{lasecki2013real}, and other semi-supervised methods~\cite{nguyen2015did,hu2016less}. Those methods typically treat HAR as a standard time series classification problem. And they tend to solve it by subsequently performing preprocessing procedures, feature extraction, model building, and activity inference. However, they all assume that the training and test data are with the same distribution. As for CDAR where the training~(source) and the test~(target) data are from different feature distributions, those conventional methods are prune to under-fitting since their generalization ability will be undermined~\cite{pan2010survey}.

Deep learning based HAR~\cite{wang2017deep} achieves the state-of-the-art performance than conventional machine learning approaches. The reason is that deep learning is capable of automatically extracting high-level features from the raw sensor readings~\cite{lecun2015deep}. Therefore, the features are likely to be more domain-invariant and tend to perform better for cross-domain tasks. A recent work evaluated deep models for cross-domain HAR~\cite{morales2016deep}, which provides some experience for future research on this area. There are stll many open problems for deep learning based CDAR. In this paper, we mainly focus on the traditional approaches.

\subsection{Transfer Learning}
Transfer learning has been successfully applied in many applications such as Wi-Fi localization~\cite{pan2008transfer}, natural language processing~\cite{blitzer2006domain}, and visual object recognition~\cite{duan2012domain}. According to the literature survey~\cite{pan2010survey}, transfer learning can be categorized into 3 types: instance-based, parameter-based, and feature-based methods. 

Instance-based methods perform knowledge transfer mainly through instance re-weighting techniques~\cite{tan2017distant,chattopadhyay2012multisource}. Parameter-based methods~\cite{yao2010boosting,zhao2011cross} first train a model using the labeled source domain, then perform clustering on the target domain. 

Our framework belongs to the feature based category, which brings the features of source and target domain into the same subspace where the data distributions can be the same. A fruitful line of work has been done in this area~\cite{wang2017balanced,long2013transfer,long2015domain}. The proposed STL  differs from existing feature-based methods in the following aspects:

\textbf{Exploit the correlations between features.} \cite{blitzer2006domain} proposed structural correspondence learning~(SCL) to generatively learn the relation of features. \cite{kouw2016feature} applied a feature-level transfer model to learn the dependence between domains, then trained a domain-adapted classifier. Instead of modeling the relationship of domain features, STL transforms the domain data into a new subspace, which does not depend on the domain knowledge in modeling features.

\textbf{Transform domains into new feature space.} \cite{pan2008transfer} proposed maximum mean discrepancy embedding~(MMDE) to learn latent features in the reproducing kernel Hilbert space~(RKHS). MMDE requires solving a semidefinite programming~(SDP) problem, which is computationally prohibitive. \cite{pan2011domain} extended MMDE by Transfer Component Analysis~(TCA), which learns a kernel in RKHS. \cite{dorri2012adapting} adopted a similar idea. \cite{seah2012learning} learns target predictive function with a low variance. \cite{glorot2011domain} sampled the domain features by viewing the data in a Grassmann manifold to obtain subspaces. \cite{gong2012geodesic} exploited the low-dimensional structure to integrate the domains according to geodesic flow kernel~(GFK). Long \textit{et al.} proposed joint distribution adaptation~(JDA) based on minimizing joint distribution between domains, while STL focuses on the marginal distribution. \cite{long2015domain} proposed transfer kernel learning~(TKL), which learned a domain-invariant kernel in RKHS. \cite{gong2016domain} studied the conditional transfer components between domains. Methods in those literature tend to learn some common representations in the new feature space, then a global domain shift can be achieved. However, the difference between individual classes is ignored.

\subsection{Transfer Learning based Activity Recognition}
Some existing work also focused on transfer learning based HAR. A detailed survey is provided in~\cite{cook2013transfer}. Among existing work, Zhao \textit{et al.} proposed a transfer learning method called TransEMDT~\cite{zhao2011cross} using decision trees, but it ignored the intra-class similarity within classes. \cite{khan2017transact} proposed the TransAct framework, which is a boosting-based method and ignores the feature transformation procedure. Thus it is not feasible in most activity cases. Feuz \textit{et al.}~\cite{feuz2017collegial} proposed a heterogeneous transfer learning method for HAR, but it only learns a global domain shift.

To circumvent the global domain shift, \cite{elhamifar2013sparse} clustered the data points in a low-dimensional space using sparse subspace clustering.  \cite{lin2016cross} identified a compact joint subspace for each class, then measured the distance between classes using principal angle. The main difference between STL and~\cite{lin2016cross} is that STL learns pseudo labels using majority voting technique on both domains, instead of performing metric clustering on the target domain alone. Thus, STL can obtain more reliable candidates by taking advantage of both domains.

\section{Stratified Transfer Learning}
\label{sec-stl}

In this section, we introduce the proposed Stratified Transfer Learning~(STL) framework. First, we present the problem definition of CDAR and the main idea of STL. Then, each step of STL is presented in detail.

\subsection{Problem Definition}
\label{sec-problem}

CDAR is a typical cross-domain learning problem, which often consists of a labeled source domain $\mathcal{D}_{s} = \{(\mathbf{x}_{i},y_{i})\}^{n_s}_{i=1}$, and an unlabeled target domain $\mathcal{D}_{t} = \{\mathbf{x}_{j}\}^{n_t}_{j=1}$. $\mathcal{D}_{s}$ and $\mathcal{D}_{t}$ have the same dimensionality and label spaces, i.e. $\mathbf{x}_{i}, \mathbf{x}_{j} \in \mathbb{R}^d$ where $d$ is the dimensionality, and $\mathcal{Y}_s = \mathcal{Y}_t$. In cross-domain learning, they follow different feature distributions. Concretely speaking, we use $P$ and $Q$ to denote the marginal and conditional distributions, respectively. Then, $P(\mathbf{x}_{s}) \neq P(\mathbf{x}_{t})$, and $Q(y_{s} | \mathbf{x}_{s}) \neq Q(y_{t}|\mathbf{x}_t)$. The goal of cross-domain learning is to obtain the labels $\mathbf{y}_t$ for the target domain $\mathcal{D}_t$ using the labeled source domain $\mathcal{D}_s$. Specifically in CDAR, the goal is to use the labeled activity information in one domain to learn the labels for another domain.

\begin{figure}[t!] 
	\centering 
	\includegraphics[scale=0.85]{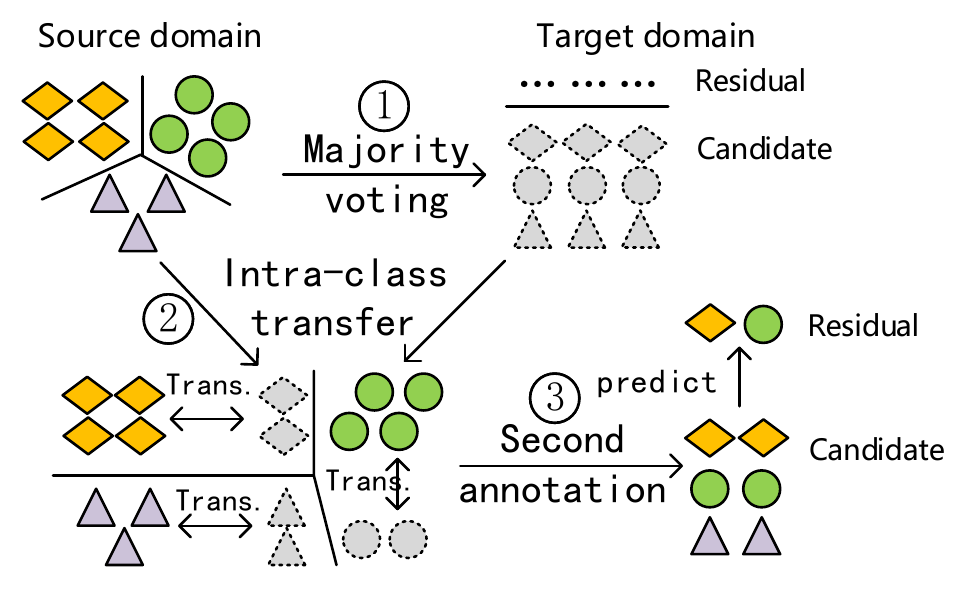}
	\vspace{-.1in}
	\caption{Main idea of stratified transfer learning~(STL) framework. There are mainly three steps: (1)~Candidates generating through majority voting technique; (2)~Perform intra-class transfer between source domain and candidates; (3)~Fully label the target domain via second annotation.}
	\label{fig-stl}
	\vspace{-.15in}
\end{figure}

STL can simultaneously transform the individual classes of the source and target domains into the same subspaces by exploiting the intra-affinity of classes. Compared to existing approaches which only learn a global domain shift, we adopt the assumption that data samples from the same class should lay on an intrinsic subspace \cite{elhamifar2013sparse}. Fig.~\ref{fig-stl} illustrates the main idea of STL. There are mainly three steps. Initially, STL generates pseudo labels for the target domain through majority voting technique. Then, it performs intra-class knowledge transfer between domains. Finally, a second annotation step is performed on the newly transformed subspaces. 

Now we elaborate each step in detail.

\subsection{Majority Voting}
As a preprocessing step of STL, this step generates pseudo labels for the target domain based on several classifiers trained on source domain. Those samples with pseudo labels are called \textit{candidates}. To this end, \textit{majority voting} technique is developed to exploit the knowledge from the crowd~\cite{prelec2017solution}. Specifically, STL makes use of some base classifiers learned on $\mathcal{D}_s$ to collaboratively learn the labels for $\mathcal{D}_t$.

Let $A_j (j = 1,2,\cdots,n_2)$ denote the final result of majority voting on $\mathbf{x}_{t_j}$, and $f_t(j)$ denotes the prediction of the $j$-th sample by the $t$-th classifier $f_t(\cdot)$, we have
\begin{equation}
\label{equ-majority}
A_j = \begin{cases}
\mathrm{majority}(\{f_t(j), t) &  \text{if majority holds}\\
-1 & \text{otherwise},
\end{cases}
\end{equation}
where $t \in \{1,2,\cdots\}$ denotes the index of classifier.

Majority voting technique generally ensembles all the classifiers learned in the source domain. The condition ``majority holds'' refers to any potential scheme that helps to generate a better solution such as simple voting and weighted voting. Specifically, $A_i$ could be defined as a) if most classifiers have same results on a sample, we take its label, else we label it `-1'; b) same as a) with voting weights to classifiers; c) the stacking of some base classifiers (similar to the deep forest proposed by Zhou and Feng~\cite{zhou2017deep}). In theory, the classifiers can be of any type in our framework.

Using majority voting, STL can generate pseudo labels for the target domain. Label `-1' means that there is no majority consensus for that sample. The samples with this label are called \textit{residuals}, and they will be annotated later. In the sequel, we will use $\mathbf{X}_{can}$ and $\mathbf{X}_{res}$ to denote the candidates and residuals, respectively. And $\widetilde{\mathbf{y}}_{can}$ denotes the pseudo labels for the candidates.

\subsection{Intra-class Transfer}
In this step, STL exploits the intra-affinity of classes to further transform each class the candidates and source domain into the same subspace. Initially, $\mathcal{D}_s$ and $\mathbf{X}_{can}$ are divided to $C$ groups according to their~(pseudo) labels, where $C$ is the total number of classes. Then, feature transformation is performed within each class of both domains. Finally, the results of distinct subspaces are merged.

The key to successful knowledge transfer is to utilize the \textit{similarity} between the source and the target domain. Hence, how to measure the similarity~(or divergence) is critical. We adopt maximum mean discrepancy~(MMD)~\cite{gretton2012kernel} as the measurement. MMD is a nonparametric method to measure the divergence between two distinct distributions and it has been widely applied to many transfer learning methods~\cite{pan2011domain,long2015domain}. The MMD distance between two domains can be formally computed as
\begin{equation}
\label{equ-mmd}
D(\mathcal{D}_{s},\mathcal{D}_{t})=
\begin{Vmatrix}
\frac{1}{n_s} \sum \limits_{\mathbf{x}_i \in \mathcal{D}_s}^{} \phi(\mathbf{x}_{i}) - \frac{1}{n_t}\sum \limits _{\mathbf{x}_j \in \mathcal{D}_t}^{} \phi(\mathbf{x}_{j})
\end{Vmatrix}^2_{\mathcal{H}}
\end{equation}
where $\mathcal{H}$ denotes reproducing kernel Hilbert space~(RKHS). Here $\phi(\cdot)$ denotes some feature map to map the original samples to RKHS. The reason we do not use the original data is that the features are often distorted in the original feature space, and it can be more efficient to perform knowledge transfer in RKHS~\cite{pan2011domain}. 

The above Eq.~\ref{equ-mmd} derives a global domain shift for the source and target domain just similar to many existing methods~\cite{pan2011domain,long2015domain}. In order to achieve \textit{intra-class transfer}, we need to calculate the MMD distance between \textit{each class}. Since the target domain has no labels, we use the pseudo labels from majority voting. For the candidates and the source domain, we present their \textit{intra-class} MMD distance as
\begin{equation}
\label{equ-stra}
\begin{split}
&\quad D(\mathcal{D}_{s},\mathbf{X}_{can})\\
&=\sum_{c=1}^{C}\left \Vert \frac{1}{n^{(c)}_s} \sum_{\mathbf{x}_{i} \in \mathcal{D}^{(c)}_s} \phi(\mathbf{x}_{i}) - \frac{1}{n^{(c)}_t} \sum_{\mathbf{x}_{j} \in \mathbf{X}^{(c)}_{can}} \phi(\mathbf{x}_{j}) \right \Vert ^2_\mathcal{H}
\end{split}
\end{equation}
where $c \in \{1, 2, \cdots, C\}$ denote the class label. $\mathcal{D}^{(c)}_s$ and $\mathbf{X}^{(c)}_{can}$ denote the samples belonging to class $c$ in the source and \textit{candidates}, respectively. $n^{(c)}_s=|\mathcal{D}^{(c)}_s|$, and $n^{(c)}_t=|\mathbf{X}^{(c)}_{can}|$.

Unfortunately, it is non-trivial to solve Eq.~(\ref{equ-stra}) directly since the mapping function $\phi(\cdot)$ is to be determined. Thus, we turn to some kernel methods. We define a kernel matrix $\mathbf{K} \in \mathbb{R}^{(n_1+n_2)\times(n_1+n_2)}$, which can be constructed by the inner product of the mapping
\begin{equation}
	\label{equ-kernel}
	\mathbf{K}_{ij}=\langle\phi(\mathbf{x}_i),\phi(\mathbf{x}_j)\rangle = \phi(\mathbf{x}_i)^\top \phi(\mathbf{x}_j)
\end{equation}
where $\mathbf{x}_i$ and $\mathbf{x}_j$ are samples from either $\mathcal{D}_s$ or $\mathbf{X}_{can}$.

We further introduce a feature transformation matrix $\mathbf{W} \in \mathbb{R}^{(n_1+n_2)\times m}$ to transform the samples of both domains from the original space to the RKHS. Here $m \ll d$ denotes the dimension after feature transformation. Then, by applying kernel tricks, Eq.~\ref{equ-stra} can be eventually formulated as the following trace optimization problem:
\begin{equation}
\label{equ-stratified}
\begin{split}
\min_{\mathbf{W}} \quad &\sum_{c=1}^{C} \mathrm{tr} (\mathbf{W}^\top \mathbf{K} \mathbf{L}_c \mathbf{K} \mathbf{W}) + \lambda \mathrm{tr}(\mathbf{W}^\top \mathbf{W})\\
\text{s.t.} \quad & \mathbf{W}^\top \mathbf{K} \mathbf{H} \mathbf{K} \mathbf{W} = \mathbf{I}
\end{split}
\end{equation}

There are two terms in the objective function of Eq.~(\ref{equ-stratified}). The first term ($\sum_{c=1}^{C} \mathrm{tr} (\mathbf{W}^\top \mathbf{K} \mathbf{L}_c \mathbf{K} \mathbf{W})$) denotes the MMD distance of each class between source and target domain, and the second one ($\lambda \mathrm{tr}(\mathbf{W}^\top \mathbf{W})$) denotes the regularization term to ensure the problem is well-defined with $\lambda$ the trade-off parameter. The constraint in Eq.~\ref{equ-stratified} is used to guarantee that the transformed data ($\mathbf{W}^\top \mathbf{K}$) will still preserve some structure property of the original data. $\mathbf{I}_{n_1+n_2}$ is the identical matrix, and $\mathbf{H} = \mathbf{I}_{n_1 + n_2} - 1/(n_1 + n_2)\mathbf{11}^\top$ is the centering matrix. For notational brevity, we will drop the subscript for $\mathbf{I}_{n_1+n_2}$ in the sequel. $\mathbf{L}_c$ is the intra-class MMD matrix, which can be constructed as:
\begin{equation}
\label{equ-lc}
(\mathbf{L}_c)_{ij}=\begin{cases}
\frac{1}{{(n^{(c)}_1)}^2} & \mathbf{x}_i,\mathbf{x}_j \in \mathcal{D}^{(c)}_s\\ 
\frac{1}{{(n^{(c)}_2)}^2} & \mathbf{x}_i,\mathbf{x}_j \in \mathbf{X}^{(c)}_{can}\\ 
-\frac{1}{n^{(c)}_1 n^{(c)}_2} & \begin{cases}
\mathbf{x}_i \in \mathcal{D}^{(c)}_s, \mathbf{x}_j \in \mathbf{X}^{(c)}_{can}\\
\mathbf{x}_i \in \mathbf{X}^{(c)}_{can}, \mathbf{x}_j \in \mathcal{D}^{(c)}_s
\end{cases}\\
0 & \text{otherwise}
\end{cases}
\end{equation}

\textbf{Learning algorithm}: Acquiring the solution of Eq.~\ref{equ-stratified} is non-trivial. To this end, we adopt Lagrange method. We denote $\bm{\Phi}$ the Lagrange multiplier, then the Lagrange function can be derived as
\begin{equation}
\label{equ-lagrange}
\begin{split}
L = ~& \mathrm{tr} \left(\mathbf{W}^\top \mathbf{K} \sum_{c=1}^{C} \mathbf{L}_c \mathbf{K}^\top \mathbf{W} \right) + \lambda \mathrm{tr}(\mathbf{W}^\top \mathbf{W})\\
& + \mathrm{tr} \left((\mathbf{I}-\mathbf{W}^\top \mathbf{K} \mathbf{H} \mathbf{K}^\top \mathbf{W})\bm{\Phi} \right)
\end{split}
\end{equation}

Setting the derivative $\partial L/\partial \mathbf{W}=0$, Eq.~\ref{equ-lagrange} can be finally formalized as an generalized eigen-decomposition problem
\begin{equation}
\label{equ-eigen}
\left(\mathbf{K} \sum_{c=1}^{C} \mathbf{L}_c \mathbf{K}^\top + \lambda \mathbf{I} \right) \mathbf{W}
=\mathbf{K} \mathbf{H} \mathbf{K}^\top \mathbf{W} \bm{\Phi}
\end{equation}

Solving Eq.~\ref{equ-eigen} refers to solve this generalized eigen-decomposition problem and take the $m$ smallest eigenvectors to construct $\mathbf{W}$. $\mathbf{W}$ can transform both domains into the same subspace with minimum domain distance while preserving their properties. Since the knowledge transfer pertains to each class, we call this step \textit{intra-class transfer}, and that is where the name~\textit{stratified} originates from. After this step, the source and target domains belonging to the same class are simultaneously transformed into the same subspaces. 

\subsection{Second annotation}
The objective of this step is to annotate the \textit{residual} part using the transformed source domain and candidates. The majority voting provides pseudo labels ($\widetilde{\mathbf{y}}_{can}$) for the \textit{candidates}, which can be transformed into the same subspaces with the target domain after intra-class transfer. Under this circumstance, it is easy to get more reliable predictions~($\hat{\mathbf{y}}_{can}$) of the \textit{candidates}. Specifically, we train a standard classifier using $\{[\mathbf{W}^\top \mathbf{K}]_{1:n_1,:},\mathbf{y}_{s}\}$ and apply prediction on $[\mathbf{W}^\top \mathbf{K}]_{n_1+1:n_2,:}$. Finally, the labels of \textit{residuals} can be obtained by training classifier on instances $\{\mathbf{X}_{can},\hat{\textbf{y}}_{can}\}$.

Since this step requires annotating the candidates again with more concrete labels, we call it \textit{second annotation}. The labels of candidates can be correspondingly close to the ground truth by annotating twice since the domains are now in the same subspace after intra-class transfer.

\subsection{Iterative refinement}
\label{sec-ite}
It should be noted that STL could achieve a better prediction if we use the result of \textit{second annotation} as the initial state and run \textit{intra-class transfer} iteratively. This \textit{EM-like} algorithm is empirically effective and will be validated in the following experiments. Additionally, we \textit{only} use majority voting in the first round of the iteration, then the proposed STL could iteratively refine the labels for the target domain by \textit{only} using its previous results.

The overall process of STL is described in Algorithm~\ref{algo-stl}. 

\textbf{Remark:} STL is a \textit{general} framework for transfer learning and it can be implemented in different ways according to the specific applications. We only provide one feasible implementation of STL in this section. Meanwhile, each step of STL can be tailored according to certain applications: 1)~majority voting could have different classifiers and voting techniques; 2)~intra-class transfer could use existing transfer learning approaches such as GFK~\cite{gong2012geodesic} and TransACT~\cite{khan2017transact}; 3)~second annotation could use different classifiers. Therefore, by considering the characteristics of certain applications, STL can be more effective and efficient in solving these problems.

\begin{algorithm}[t!]
	\caption{STL:~\underline{S}tratified \underline{T}ransfer \underline{L}earning}
	\label{algo-stl}
	\renewcommand{\algorithmicrequire}{\textbf{Input:}} 
	\renewcommand{\algorithmicensure}{\textbf{Output:}} 
	\begin{algorithmic}[1]
		\REQUIRE~~
		Source domain $\mathcal{D}_{s}=\{\mathbf{X}_{s},\mathbf{y}_{s}\}$, target domain $\mathcal{D}_{t}=\{\mathbf{X}_{t}\}$, \#dimension $m$.\\
		\ENSURE~~
		The labels for the target domain: $\{\mathbf{y}_{t}\}$.\\
		\STATE Perform majority voting on $\mathcal{D}_{t}$ using Eq.~\ref{equ-majority} to get $\{\mathbf{X}_{can},\widetilde{\mathbf{y}}_{can}\}$ and $\mathbf{X}_{res}$;
		\STATE Construct kernel matrix $\mathbf{K}$ according to Eq.~(\ref{equ-kernel}) using $\mathbf{X}_{src}$ and $\mathbf{X}_{can}$, and compute the intra-class MMD matrix $\mathbf{L}_c$ using Eq.~\ref{equ-lc};
		\REPEAT
		\STATE Solve the eigen-decomposition problem in Eq.~\ref{equ-eigen} and take the $m$ smallest eigen-vectors to obtain the transformation matrix $\mathbf{W}$;
		\STATE Transform the same classes of $\mathbf{X}_s$ and $\mathbf{X}_{can}$ into the same subspaces using $\mathbf{W}$, and then merge them;
		\STATE Perform second annotation to get $\{\hat{\mathbf{y}}_{can}\}$ and $\{\hat{\mathbf{y}}_{res}\}$;
		\STATE Construct kernel matrix $\mathbf{K}$ and compute the intra-class MMD matrix $\mathbf{L}_c$ using Eq.~\ref{equ-lc};
		\UNTIL{Convergence}
		\RETURN $\{\mathbf{y}_{t}\}$.
	\end{algorithmic}
\end{algorithm}

\section{Experimental Evaluation}
\label{sec-exp}
In this section, we evaluate the performance of STL via extensive experiments on cross-domain activity recognition.

\begin{table*}[t!]
	\centering
	\caption{Brief introduction of three public datasets for activity recognition}
	\vspace{-.1in}
	\label{tb-dataset}
	\resizebox{\textwidth}{!}{
		\begin{tabular}{|c|c|c|c|c|c|}
			\hline
			\textbf{Dataset} & \textbf{\#Subject} & \textbf{\#Activity} & \textbf{\#Sample} & \textbf{\#Feature} & \textbf{Body parts} \\ \hline \hline
			OPPORTUNITY & 4 & 4 & 701,366 & 459 & Back (B), Right Upper Arm (RUA), Right Left Arm (RLA), Left Upper Arm (LLA), Left Lower Arm (LLA) \\ \hline
			PAMAP2 & 9 & 18 & 2,844,868 & 243 & Hand (H), Chest(C), Ankle (A) \\ \hline
			DSADS & 8 & 19 & 1,140,000 & 405 & Torso (T), Right Arm (RA), Left Arm (LA), Right Leg (RL), Left Leg (LL) \\ \hline
		\end{tabular}
	}
	\vspace{-.1in}
\end{table*}

\subsection{Cross-domain activity recognition}
Cross-domain activity recognition~(CDAR) aims at labeling the activity of one domain using the labeled data from another related domain. There are several types of CDAR: cross-person, cross-device, cross-environment activity recognition, and so on. In our experiments, we choose \textit{cross-position activity recognition}. Specifically, it refers to the situation where the activity labels of some body parts are missing, so it is necessary and feasible to leverage the labeled data from other similar body parts to get the labels of those body parts. 

\textit{Why cross-position activity recognition}: We need to extensively investigate the performance of transfer learning at different degrees of similarities between the source and target domain. In cross-position tasks, there are different degrees of similarities, since body parts are different at certain degrees. For instance, there is more similarity between both arms than between arm and torso. Compared to other types of CDAR such as cross-person or cross-device, cross-position tasks can provide more detailed information about domain similarities, thus more experience about activity transfer can be revealed.

\textit{Why knowledge can be transferred between body parts}: First of all, different body parts tend to share similar structure and functions. Then, the activity patterns of different body parts are often related. Thus, annotating the activity becomes a cross-domain learning problem because the data distribution is different on body parts.

\subsection{Datasets and Preprocessing}
Three large public datasets are adopted in our experiments. TABLE~\ref{tb-dataset} provides a brief introduction to those three datasets. In the following, we briefly introduce those datasets, and more information can be found in their original papers. OPPORTUNITY dataset~(\textbf{OPP})~\cite{chavarriaga2013opportunity} is composed of 4 subjects executing different levels of activities with sensors tied to more than 5 body parts. PAMAP2 dataset~(\textbf{PAMAP2})~\cite{reiss2012introducing} is collected by 9 subjects performing 18 activities with sensors on 3 body parts. UCI daily and sports dataset~(\textbf{DSADS})~\cite{barshan2014recognizing} consists of 19 activities collected from 8 subjects wearing body-worn sensors on 5 body parts. Accelerometer, gyroscope, and magnetometer are all used in three datasets.

\begin{figure}
	\centering
	\includegraphics[scale=0.48]{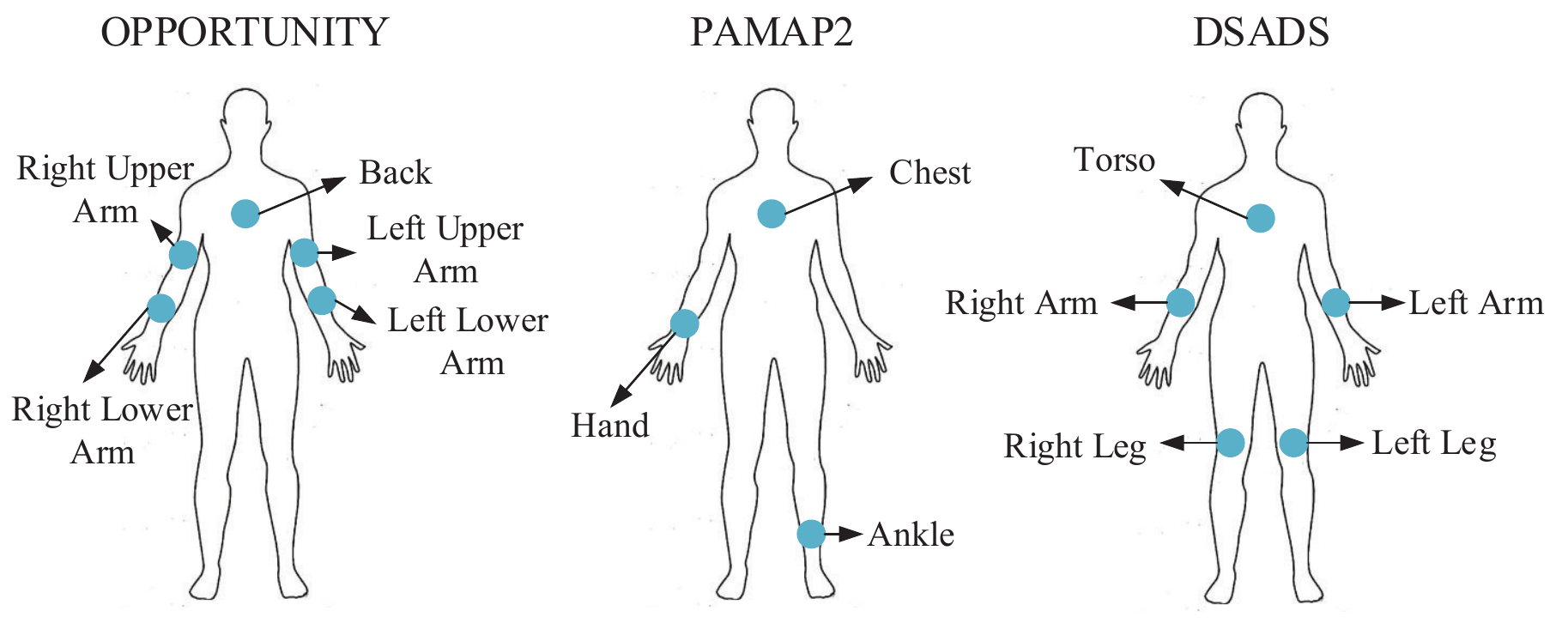}
	\vspace{-.2in}
	\caption{Different positions on OPPOPTUNITY, PAMAP2 and DSADS.}
	\label{fig-dataset}
	\vspace{-.1in}
\end{figure}

Fig.~\ref{fig-dataset} illustrates the positions we investigated in three datasets. In our experiments, we use the data from all three sensors in each body part since most information can be retained in this way. For one sensor, we combine the data from 3 axes together using $a=\sqrt{x^2+y^2+z^2}$. Then, we exploit the sliding window technique to extract features~(window length is 5s). The feature extraction procedure is mainly executed according to existing work~\cite{hu2017okrelm}. In total, 27 features from both time and frequency domains are extracted for a single sensor~(see TABLE~\ref{tb-feature}). Since there are three sensors~(i.e. accelerometer, gyroscope, and magnetometer) on one body part, we extracted 81 features from one position.

In order to investigate more detailed information during activity knowledge transfer, we perform CDAR in the following three scenarios according to the different similarities between body parts. Those three scenarios represent three different degrees of similarities between domains. In the sequel, we use the notation $A \rightarrow B$ to denote labeling the activity of domain $B$ using the labeled domain $A$.

\begin{itemize}[noitemsep,nolistsep]
	\item a)~\textit{similar body parts of the same person}, which refers to highly similar body parts such as left and right arms on the same person. Specifically, we extracted the data from OPP and DSADS~(PAMAP2 is not included since there are no similar parts), and constructed 8 tasks: RA $\rightarrow$ LA, RL $\rightarrow$ LL, RUA $\rightarrow$ LUA, RLA $\rightarrow$ LLA, and vice versa.
	\item b)~\textit{different body parts of the same person}, which refers to different body parts like torso and arm of the same person. Specifically, we constructed 8 tasks from three datasets: RA $\rightarrow$ T, H $\rightarrow$ C, RLA $\rightarrow$ T, RUA $\rightarrow$ T, and vice versa.
	\item c)~\textit{similar body parts of different person}, which refers to activity recognition on same body parts across different datasets. Specifically, we constructed 6 tasks for T $\rightarrow$ T across three datasets~\footnote{In this paper, for brevity, we use ``T'' to denote the Torso / Back / Chest in three datasets, respectively.}.
\end{itemize}

In total, we constructed 22 tasks. Note that there are different activities in three datasets. For scenario a) and b), we simply use all the classes in each dataset; for scenario c)~which is cross-dataset, we extract 4 common classes for each dataset~(i.e. \textit{Walking, Sitting, Lying}, and \textit{Standing}). In addition, we did not include the scenario `\textit{different body parts of different person}' since 1)~all the methods perform poorly in that scenario, and 2)~that scenario does not have reasonable feasibility in real applications.

\begin{table}[!t]
	\centering
	\caption{Features extracted per sensor on each body part}
	\vspace{-.1in}
	\label{tb-feature}
	\resizebox{0.48\textwidth}{!}{
		\begin{tabular}{@{}lll@{}}
			\toprule
			\textbf{ID}    & \textbf{Feature}                 & \textbf{Description}                                  \\ \midrule
			1     & Mean                    & Average value of samples in window           \\ 
			2     & STD                     & Standard deviation                           \\ 
			3     & Minimum                 & Minimum                                      \\ 
			4     & Maximum                 & Maximum                                      \\ 
			5     & Mode                    & The value with the largest frequency         \\
			6     & Range                   & Maximum minus minimum                        \\ 
			7     & Mean crossing rate      & Rate of times signal crossing mean value\\ 
			8     & DC                      & Direct component                             \\ 
			9-13  & Spectrum peak position  & First 5 peaks after FFT                      \\ 
			14-18 & Frequency               & Frequencies corresponding to 5 peaks\\ 
			19    & Energy                  & Square of norm                               \\ 
			20-23 & Four shape features     & Mean, STD, skewness, kurtosis                \\ 
			24-27 & Four amplitude features & Mean, STD, skewness, kurtosis                \\ \bottomrule
		\end{tabular}
	}
\vspace{-.1in}
\end{table}

\begin{table*}[ht!]
	\centering
	\caption{Classification accuracy (\%) of STL and other comparison methods on different CDAR tasks.}
	\label{tb-results}
	\vspace{-.1in}
	\resizebox{1.0\textwidth}{!}{
	\begin{tabular}{|c|c|c|c|c|c|c|c|c|}
		\hline
		Scenario & Dataset & \multicolumn{1}{c|}{Task} & \multicolumn{1}{c|}{PCA} & \multicolumn{1}{c|}{KPCA} & \multicolumn{1}{c|}{TCA} & \multicolumn{1}{c|}{GFK} & \multicolumn{1}{c|}{TKL} & \multicolumn{1}{c|}{STL} \\ \hline \hline
		\multirow{4}{*}{Similar body parts on same person} & \multirow{2}{*}{DSADS} & RA $\rightarrow$ LA & 59.91 & 62.17 & 66.15 & \textbf{71.07} & 54.10 & \textbf{71.04} \\ \cline{3-9} 
		&  & RL $\rightarrow$ LL & 69.46 & 70.92 & 75.06 & 79.71 & 61.63 & \textbf{81.60} \\ \cline{2-9} 
		& \multirow{2}{*}{OPP} & RUA $\rightarrow$ LUA & 76.12 & 65.64 & 76.88 & 74.62 & 66.81 & \textbf{83.96} \\ \cline{3-9} 
		&  & RLA $\rightarrow$ LLA & 62.17 & 66.48 & 60.60 & 74.62 & 66.82 & \textbf{83.93} \\ \hline
		\multirow{4}{*}{Different body parts on same person} & DSADS & RA $\rightarrow$ T & 38.89 & 30.20 & 39.41 & 44.19 & 32.72 & \textbf{45.61} \\ \cline{2-9} 
		& PAMAP2 & H $\rightarrow$ C & 34.97 & 24.44 & 34.86 & 36.24 & 35.67 & \textbf{43.47} \\ \cline{2-9} 
		& \multirow{2}{*}{OPP} & RLA $\rightarrow$ T & \textbf{59.10} & 46.99 & 55.43 & 48.89 & 47.66 & 56.88 \\ \cline{3-9} 
		&  & RUA $\rightarrow$ T & 67.95 & 54.52 & 67.50 & 66.14 & 60.49 & \textbf{75.15} \\ \hline
		\multirow{3}{*}{Similar body parts on different person} & PAMAP2 $\rightarrow$ OPP & T $\rightarrow$ T & 32.80 & \textbf{43.78} & 39.02 & 27.64 & 35.64 & 40.10 \\ \cline{2-9} 
		& DSADS $\rightarrow$ PAMAP & T $\rightarrow$ T & 23.19 & 17.95 & 23.66 & 19.39 & 21.65 & \textbf{37.83} \\ \cline{2-9} 
		& OPP $\rightarrow$ DSADS & T $\rightarrow$ T & 44.30 & 49.35 & 46.91 & 48.07 & 52.79 & \textbf{55.45} \\ \hline \hline
		\multicolumn{3}{|c|}{Average} & 51.71 & 48.40 & 53.23 & 53.69 & 48.73 & \textbf{61.37} \\ \hline
	\end{tabular}
}
\vspace{-.15in}
\end{table*}

\subsection{Comparison Methods and Implementation Details}
We adopt five state-of-the-art comparison methods:
\begin{itemize}
	\item PCA: Principal component analysis~\cite{fodor2002survey}.
	\item KPCA: Kernel principal component analysis~\cite{fodor2002survey}.
	\item TCA: Transfer component analysis~\cite{pan2011domain}.
	\item GFK: Geodesic flow kernel~\cite{gong2012geodesic}.
	\item TKL: Transfer kernel learning~\cite{long2015domain}.
\end{itemize}

PCA and KPCA are classic dimensionality reduction methods, while TCA, GFK, and TKL are representative transfer learning approaches. The codes of PCA and KPCA are provided in Matlab. The codes of TCA, GFK, and TKL can be obtained online \footnote{\url{https://tinyurl.com/y79j6twy}}. The constructed datasets and STL code are released online~\footnote{\url{https://tinyurl.com/y7en6owt}}.

We construct the tasks according to each scenario and use the labels for target domain only for testing. Then we perform CDAR using STL and all comparison methods. Other than TKL, all other methods require dimensionality reduction. therefore, they were tested using the same dimension. After that, a classifier with the same parameter is learned using the source domain and then the target domain can be labeled. To be more specific, we use the random forest classifier~($\#Tree=30$) as the final classifier for all the 6 methods. For majority voting in STL, we simply use SVM~($C=100$), kNN~($k=3$), and random forest~($\#Tree=30$) as the base classifiers. Other parameters are searched to achieve their optimal performance. The \#iteration is set to be $T=10$ for STL. It is noticeable that we randomly shuffle the experimental data for 5 times in order to gain robust results.

Classification $accuracy$ on the target domain is adopted as the evaluation metric, which is widely used in existing transfer learning methods~\cite{long2015domain,pan2011domain}
\begin{equation}
Accuracy = \frac{|\mathbf{x}: \mathbf{x} \in \mathcal{D}_t \wedge \hat{y}(\mathbf{x})=y(\mathbf{x})|}{\mathbf{x}:\mathbf{x} \in \mathcal{D}_t}
\end{equation}
where $y(\mathbf{x})$ and $\hat{y}(\mathbf{x})$ are the truth and predicted labels, respectively.

\subsection{Classification Accuracy of STL}
\label{sec-exp-performance}

We run STL and other methods on all CDAR tasks and report the classification accuracy in TABLE~\ref{tb-results}. For brevity, we only report the results of $A \rightarrow B$  since its result is close to $B \rightarrow A$. It is obvious that STL significantly outperforms other methods in most cases~(with a remarkable improvement of $\mathbf{7.68}\%$ over the best baseline GFK). Compared to traditional dimensionality reduction methods~(PCA and KPCA), STL improves the accuracy by $10\% \sim 20\%$, which implies that STL is better than typical dimensionality reduction methods. Compared to transfer learning methods~(TCA, GFK, and TKL), STL still shows an improvement of $5\% \sim 15\%$. Therefore, STL is more effective than all the comparison methods in most cases.

The performance of TKL is the worst, because of the instability of the transfer kernel. TCA only learns a global domain shift, thus the similarity within classes is not fully exploited. The performance of GFK is second to STL, even if GFK also learns a global domain shift. Because the geodesic distance in high-dimensional space is capable of preserving the intra properties of domains.

The differences between STL and GFK are: 1)~STL outperforms GFK in most cases with significant improvement; 2)~STL strongly outperforms GFK in \textit{less} similar scenarios~(scenarios~a) and~b)), indicating that STL is more robust in recognizing different levels of activities. For STL, it performs intra-class knowledge transfer after generating pseudo labels for candidates. Thus, better performance can be achieved by exploiting the intra-affinity of classes.

\begin{figure*}[t!]
	\centering
	\subfigure[Different degrees of similarities]{
		\centering
		\includegraphics[scale=0.45]{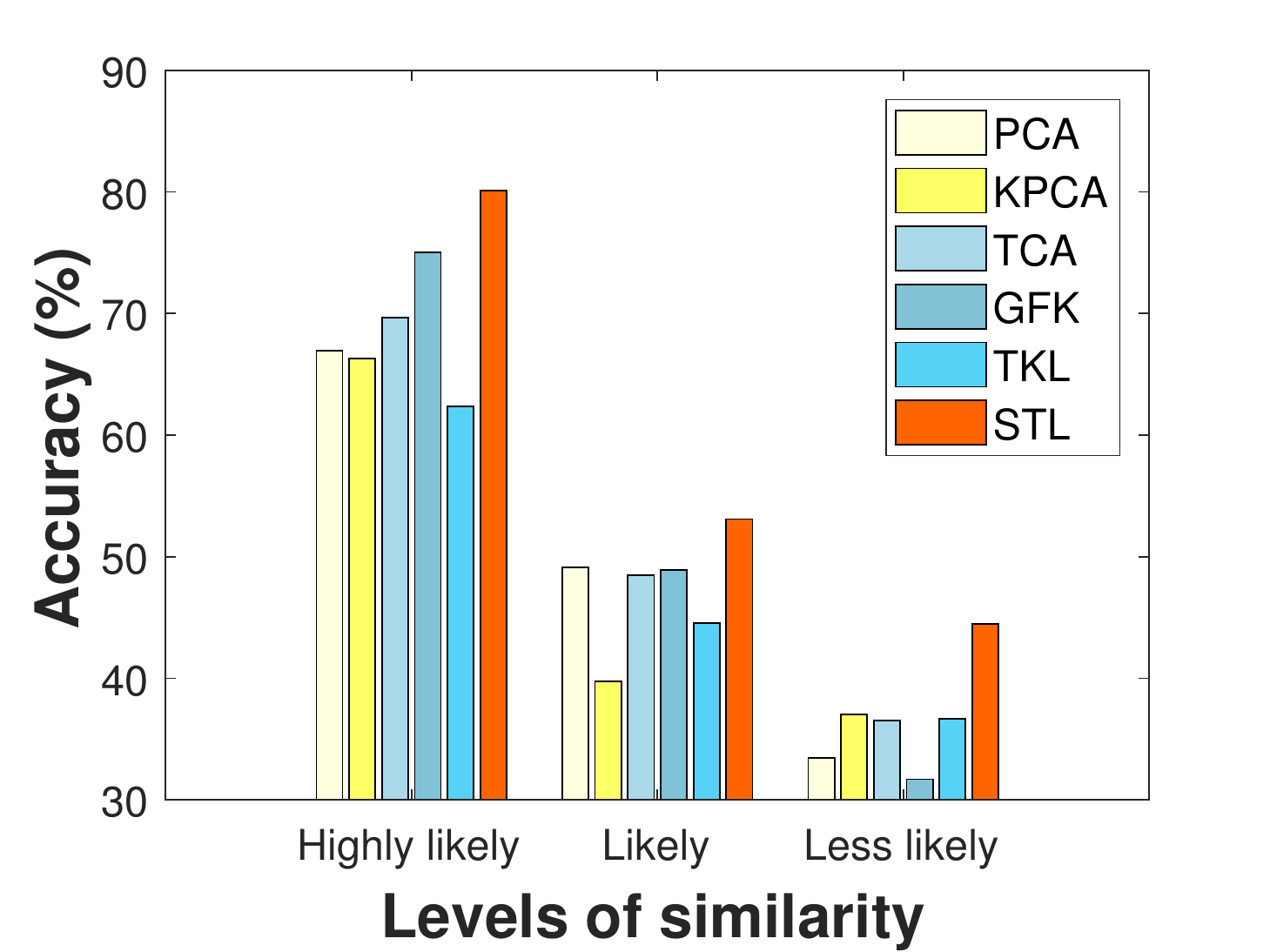}
		\label{fig-sub-sim}}
	\hspace{2cm}
	\subfigure[Different levels of activities of 3 persons]{
		\centering
		\includegraphics[scale=0.45]{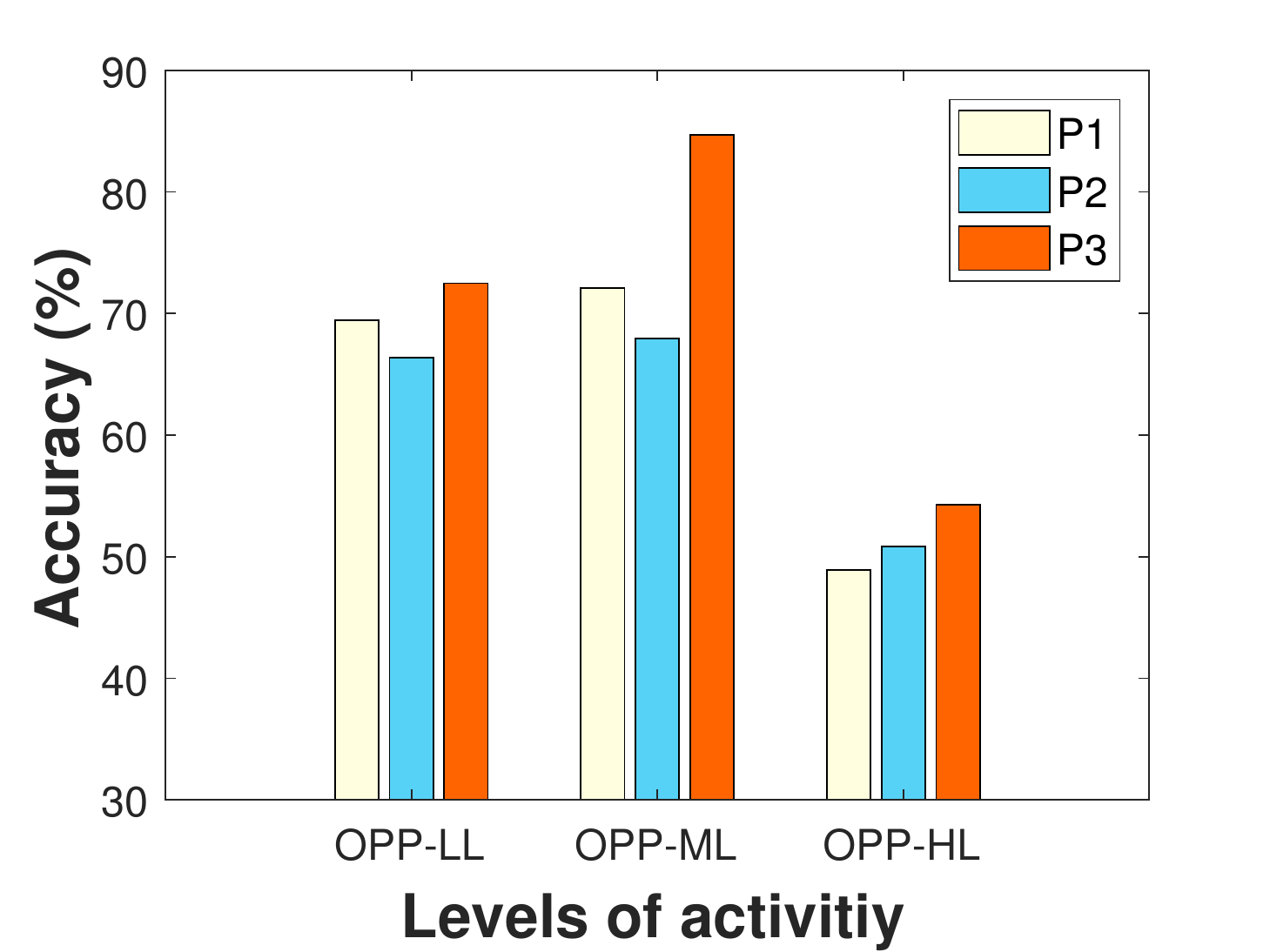}
		\label{fig-sub-level}}
	\label{fig-sim}
	\vspace{-.1in}
	\caption{Classification accuracy of~(a)~different degrees of similarities and~(b)~different levels of activities in transfer learning.}
	\vspace{-.2in}
\end{figure*}

\subsection{Performance on Different Degrees of Similarities}
\label{sec-exp-sim}
We explore the performance of the transfer learning methods under different degrees of similarities between the source and target domain. The average classification accuracy of each method is shown in Fig.~\ref{fig-sub-sim}. For all the methods, the accuracy drops as the domain similarity becomes less. Additionally, the performance of STL is the best in all scenarios.

More elaborate observations can be given combining TABLE~\ref{tb-results} and Fig.~\ref{fig-sub-sim}. In different scenarios, the change of classification accuracy of each method follows the same tendency. Specifically, the performance of all the methods is the best between similar body parts for the same person~(e.g. RA $\rightarrow$ LA). The performance becomes worse for different body parts~(e.g. RA $\rightarrow$ T). This is because Right Arm~(RA) is more similar to Left Arm~(LA) than to Torso~(T). For a different person, all methods produce the worst results because different people have exactly different body structure and moving patterns~(e.g.~T $\rightarrow$ T across datasets).

At the same degree of similarity, the results are also different. For example, the performance of RUA $\rightarrow$ T is better than RLA $\rightarrow$ T in the same dataset. Because there is more similarity between Right Upper Arm~(RUA) and Torso~(T) than between Right Lower Arm~(RLA) and Torso~(T). Other positions also share the similar discovery.

All the experimental results indicate that the \textit{similarity} between the source and target domain is important for cross-domain learning. It is critical to find the \textbf{right} auxiliary domain to perform successful knowledge transfer. In our CDAR experiments, the body structure and moving patterns contribute to the similarity. In real life, other factors such as age and hobby also help define the similarity of activities. For other cross-domain tasks~(image classification etc.), finding the relevant domain is also important. For example, the most similar image set for a dog is probably a cat, since those two kinds of animals have similar body structures, moving patterns, and living environments.

\subsection{Performance on Different Levels of Activities}
\label{sec-exp-activity}
Different levels of activities imply the different depths of activity granularities. In this section, we extensively investigate the performance of our proposed STL on CDAR tasks at different activity levels. By taking advantage of the diverse activity classes in OPP dataset~\cite{chavarriaga2013opportunity}, we analyze the transfer learning performance on low-level~(\textbf{OPP-LL}), middle-level~(\textbf{OPP-ML}), and high-level~(\textbf{OPP-HL}) activities. The results are presented in Fig.~\ref{fig-sub-level}. It indicates the best performance is achieved at \textit{middle-level} activities, while it suffers from low-level and even worse at high-level activities.

Low-level activities such as \textit{Walking} and middle-level activities such as \textit{Closing} are mostly contributed by the atomic movements of body parts and they are likely to achieve better transfer results than the high-levels. On the other hand, high-level activities such as \textit{Coffee Time} not only involve basic body movements, but also contain \textit{contextual} information like ambient or objects, which is difficult to capture only by the body parts. Since the bridge of successful cross-position transfer learning is the similarity of body parts, it is not ideal to achieve good transfer performance by only using body parts. The reason why results on OPP-ML are better than OPP-LL is that activities of OPP-ML are more fine-grained than OPP-LL, making it more capable of capturing the similarities between the body parts.

\begin{figure*}[t!]
	\centering
	\subfigure[Candidate usage]{
		\centering
		\includegraphics[scale=0.38]{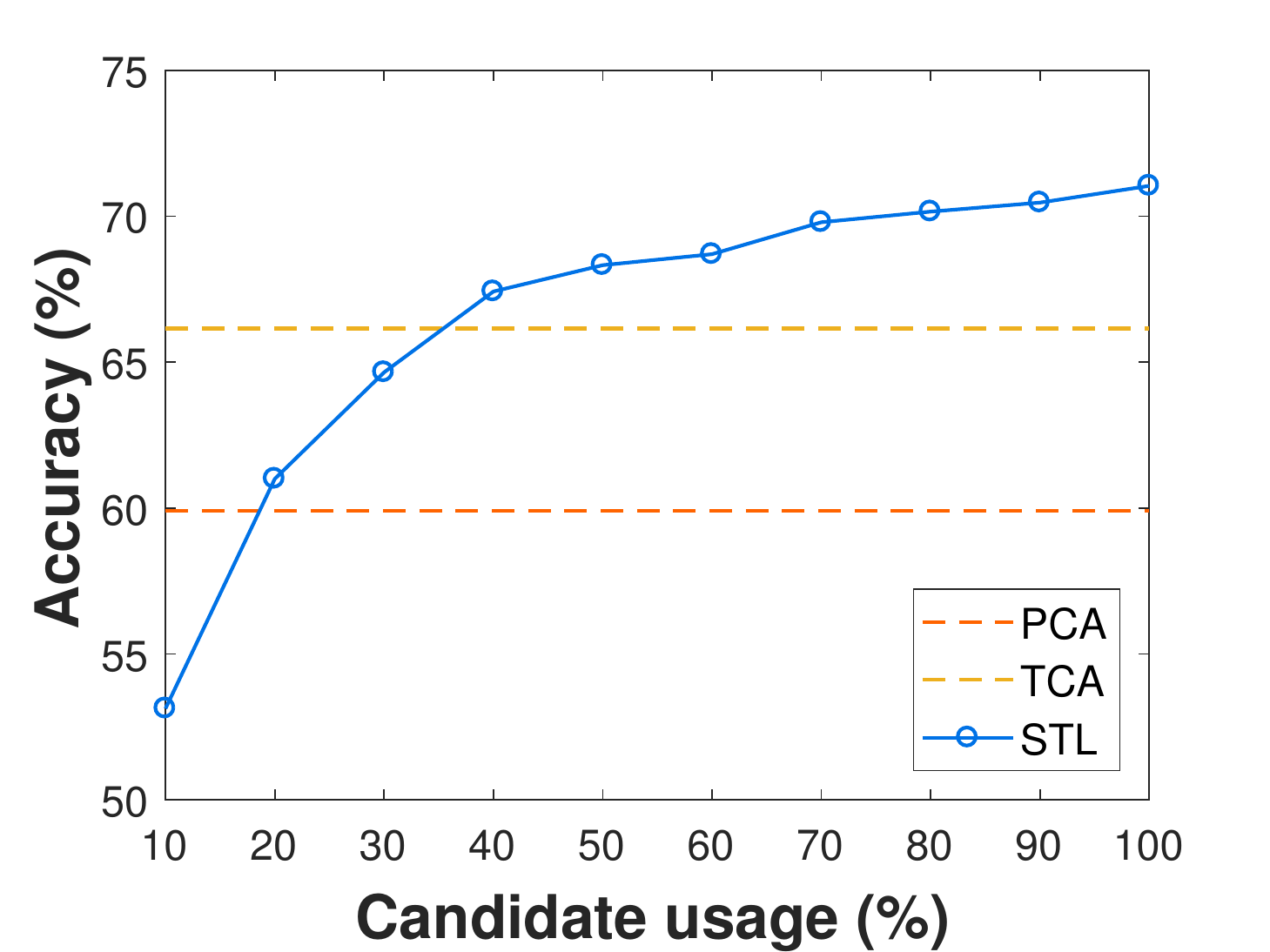}
		\label{fig-sub-candidate}}
	\subfigure[Iteration]{
		\centering
		\includegraphics[scale=0.38]{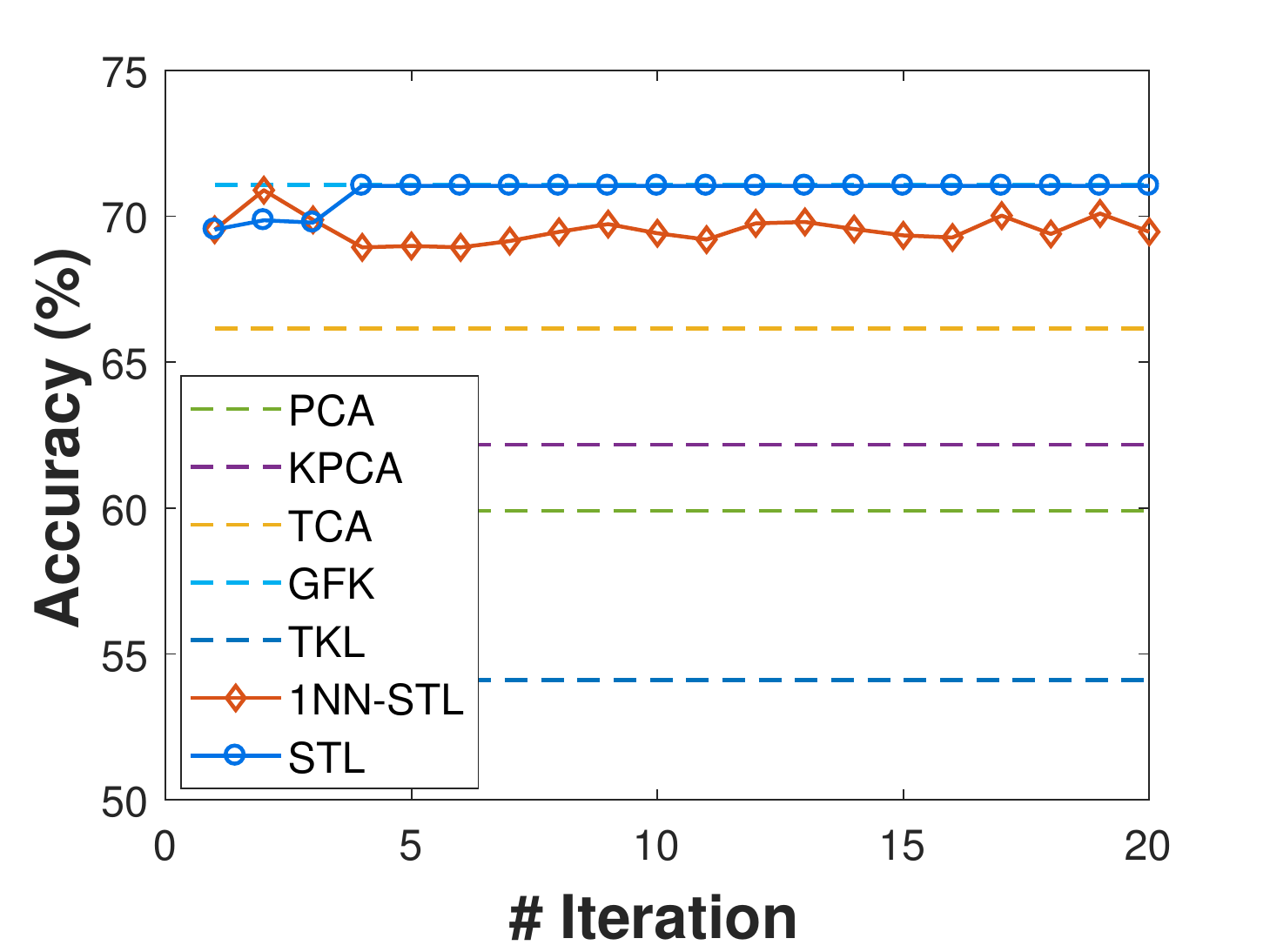}
		\label{fig-sub-ite}}
	\subfigure[Parameter sensitivity]{
		\centering
		\includegraphics[scale=0.38]{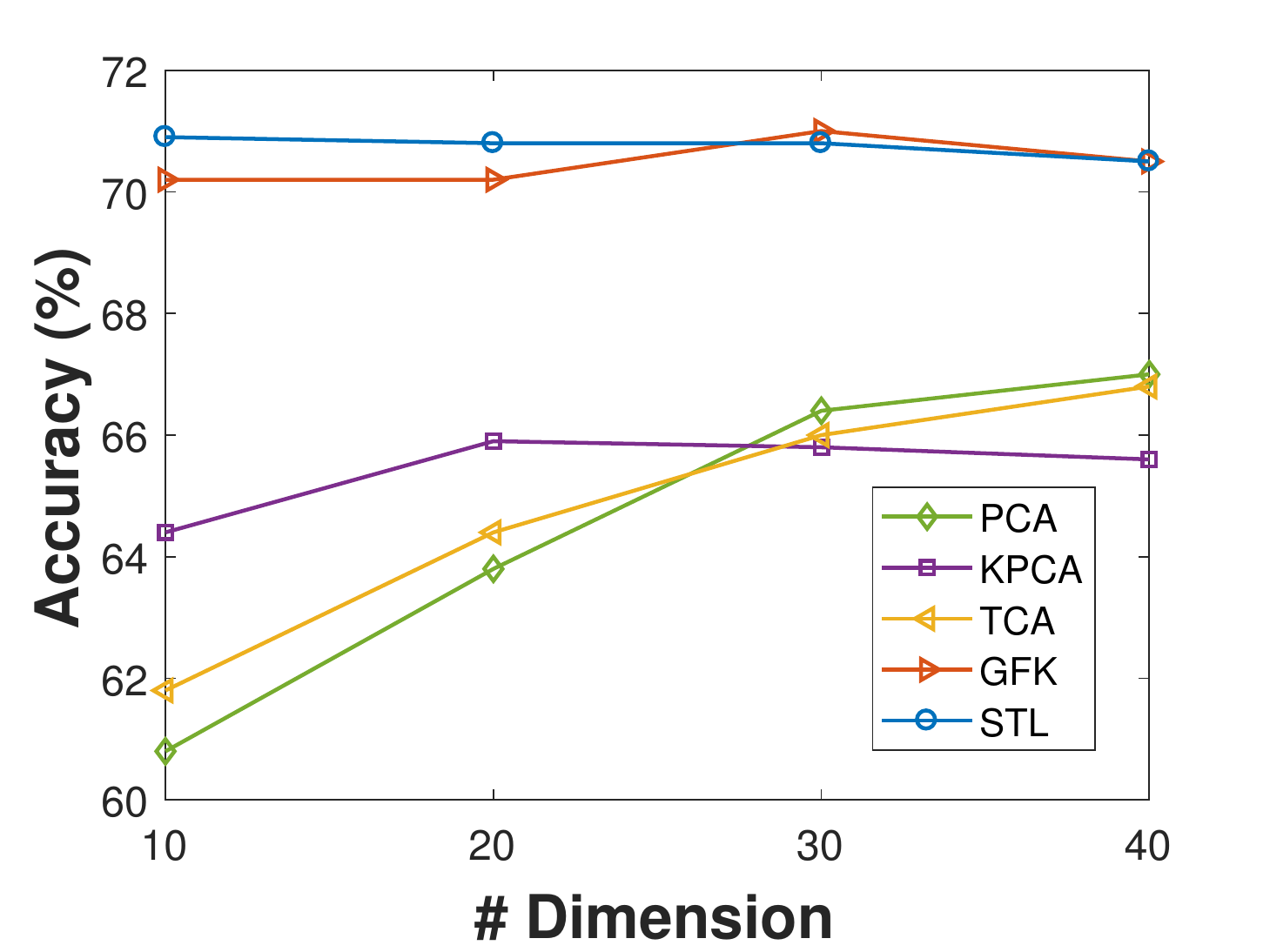}
		\label{fig-sub-para}}
	\label{fig-last}
	\vspace{-.1in}
	\caption{Detailed results of a) STL with increasing usage of candidates for RA $\rightarrow$ LA on DSADS dataset; b) 1NN-STL and STL with other methods; c) parameter sensitivity of STL in comparison with other methods}
	\vspace{-.1in}
\end{figure*}

\subsection{Effectiveness verification of STL}
In this section, we verify the effectiveness of STL in several aspects. The core idea of STL is \textit{intra-class transfer}, where the \textit{pseudo} labels of the \textit{candidates} are acting as the evidence of transfer learning. It is intuitive to ask the following three questions. Firstly, \textit{Can the confidence of the pseudo labels have an influence on the algorithm?} It seems that STL will not achieve good performance if there is not enough \textit{reliable} candidates available. Secondly, \textit{Can the choice of majority voting classifiers affect the framework?} If we use different classifiers, the performance is likely to vary. Thirdly, \textit{Can intra-class transfer be implemented in parallel?} Since in real applications, there are often several classes, which would require a large amount of training time if not in parallel. In this part, we answer those questions through the following experiments. 

\textbf{1) The confidence of the candidates:} We control the percentage of the \textit{candidates} from 10\% to 100\% in every trial and make the rest belong to the \textit{residual} part. Then we test the performance of STL. We test on the task LA $\rightarrow$ RA on DSADS and compare with other 2 methods (PCA and TCA). The result is shown in Fig.~\ref{fig-sub-candidate}. For simplicity, we only compare STL with PCA and TCA, since they are both classic dimensionality reduction methods. From the results, we can observe that the performance of STL is increasing along with the increment of candidates percentage. More importantly, STL outperforms the other two methods with \textbf{less than} 40\% of candidates. It reveals that STL does not largely rely on the confidence of the candidates and can achieve good performance even with fewer candidates.

\textbf{2) Majority voting classifiers and iteration:} To test the effectiveness of majority voting classifiers, we choose 1 nearest neighbor~(1NN) as the base classifier of majority voting. Then we run STL iteratively. The results are shown in Fig.~\ref{fig-sub-ite}, where `1NN-STL' and `STL' denotes the result of STL using 1NN and random forest as the base majority voting classifier, respectively. From those results, we can observe: 1) STL can iteratively improve the classification accuracy even with some weak majority voting classifier. 1NN-STL achieves slightly worse than STL, indicating that STL is rather \textit{robust} to the base classifiers. Since more powerful classifiers would lead to better performance, we strongly suggest using more reliable classifiers for majority voting in real problems. 2) STL can reach a quick convergence within \textit{fewer than} 10 iterations. This indicates that STL can be efficiently trained.

\textbf{3) Potential of parallel deployment:} STL has another significant advantage: it can easily be implemented in \textit{parallel}, which is rather important in real applications. Two steps of STL intuitively have that excellence: 1)~majority voting involves more than one classifier, where each classifier can be trained and used individually; 2)~the intra-class transfer can be performed within each class separately. Those two properties demonstrate that STL could be more powerful and efficient if deployed in parallel in real applications. We will continue to explore that advantage in the future.

\subsection{Parameter Sensitivity}
\label{sec-exp-dim}
STL involves two parameters: the dimension $m$, and trade-off parameter $\lambda$. In this experiment, we empirically evaluate the sensitivity of $m$. The evaluation of $\lambda$ is omitted due to page limit and our experiments have verified its robustness.

We set $m \in \{10, 20, 30, 40\}$ and test the performance of STL and other dimensionality reduction methods. As shown in Fig.~\ref{fig-sub-para}, STL achieves the best accuracy under different dimensions. Meanwhile, the accuracy of STL almost does not change with the decrement of $m$. The results reveal that STL is much more effective and robust than other methods under different dimensions. Therefore, STL can be easily applied to many cross-domain tasks which require robust performance w.r.t. different dimensions.

\section{Discussion}
\label{sec-dis}
We studied cross-domain activity recognition via the STL framework and evaluated its performance on cross-position HAR tasks. There are more applications in pervasive computing that STL could be applied to. In this section, we discuss the potential of STL in other applications and give some empirical suggestions.

\textit{1) Activity recognition.} The results of activity recognition can be different according to different \textit{devices}, \textit{users}, and wearing \textit{positions}. STL makes it possible to perform cross-device/user/position activity recognition with high accuracy. In case cross-domain learning is needed, finding and measuring the similarity between the device/user/position is critical.

\textit{2) Localization.} In WiFi localization, the WiFi signal changes with the \textit{time}, \textit{sensor}, and \textit{environment}, causing the distributions different. So it is necessary to perform cross-domain localization. When applying STL to this situation, it is also important to capture the similarity of signals according to time/sensor/environment.

\textit{3) Gesture recognition.} For gesture recognition, due to differences in hand structure and moving patterns, the model cannot generalize well. In this case, STL can also be a good option. Meanwhile, special attention needs to be paid to the divergence between the different characteristic of the subjects. 

\textit{4) Other context-related applications.} Other applications include smart home sensing, intelligent city planning, healthcare, and human-computer interaction. They are also context-related applications. Most of the models built for pervasive computing are only \textit{specific} to certain contexts. Transfer learning makes it possible to transfer the knowledge between related contexts, of which STL can achieve the best performance. But when recognizing high-level contexts such as \textit{Coffee Time}, it is rather important to consider the relationship between different contexts in order to utilize their similarities. The research on this area is still on the go.

\textit{5) Suggestions for using STL.} Firstly, before using STL and other transfer learning approaches, it is critical to investigate the similarity between the source and target domain in advance. Secondly, it is better to use rather strong classifiers for majority voting since it makes the convergence quick. Thirdly, STL can be more efficient if deployed in parallel. Additionally, each step of STL can be tailored according to specific applications.

\section{Conclusion and Future Work}
\label{sec-conclu}
The label scarcity problem is very common in pervasive computing. Transfer learning addresses this issue by leveraging labeled data from auxiliary domains to annotate the target domain. In this paper, we propose a novel and general Stratified Transfer Learning~(STL) framework for cross-domain learning in pervasive computing. Compared to existing approaches which only obtain a global domain shift, STL can exploit the intra-affinity of each class between different domains. We also provide a solution to implement STL. Experiments on three large public datasets demonstrate the significant superiority of STL over existing five state-of-the-art methods. In addition, we extensively analyze the performance of transfer learning under different degrees of similarities and different levels of activities. We also extensively discuss the potential of STL in other pervasive computing applications. In addition, STL is a general framework where each step can be tailored according to specific applications, making STL more effective and efficient in solving other problems.

In the future, we plan to extend STL using deep learning, as well as evaluating STL through more cross-domain learning problems in pervasive computing.

\section*{Acknowledgments}
J. Wang would like to thank Dr. Shuji Hao from A*star, Singapore, and Dr. Mingsheng Long from THU, China, for their valuable comments. This work is supported in part by National Key R \& D Plan of China~(No.2016YFB1001200), NSFC~(No.61572471), and Science and Technology Planning Project of Guangdong Province~(No.2015B010105001).

\bibliographystyle{IEEEtran}
\bibliography{percom18}

\end{document}